\documentclass{article}

     \usepackage[preprint]{neurips_2019}

\usepackage[utf8]{inputenc} 
\usepackage[T1]{fontenc}    
\usepackage{hyperref}       
\usepackage{url}            
\usepackage{booktabs}       
\usepackage{amsfonts}       
\usepackage{nicefrac}       
\usepackage{microtype}      

\usepackage{csvsimple}      
\usepackage{algorithm}      
\usepackage{algpseudocode}  
\usepackage{amsmath}        
\usepackage{graphicx}       
\usepackage{subcaption}     
\usepackage{multirow}       

\title{Embedded hyper-parameter tuning \\by Simulated Annealing}

\author{
  Matteo Fischetti \thanks{corresponding author: \texttt{http://www.dei.unipd.it/\textasciitilde fisch}} \\
  Department of Information Engineering\\
  University of Padova, Italy\\
  \texttt{matteo.fischetti@unipd.it} \\
  \And
  Matteo Stringher \\
  Department of Information Engineering\\
  University of Padova, Italy\\
  \texttt{stringher.matteo@gmail.com} 
}

\def\NOI{{\tt SGD-SA}}    
\def\SGD{{\tt Scheduled-SGD}}    
\def\SSA{{\tt SSA}}    
      
\def\mathbf{}       

\newif\iflong     

\begin{document}

\maketitle

\begin{abstract}

We propose a new metaheuristic training scheme that combines Stochastic Gradient Descent (SGD) and Discrete Optimization in an unconventional way.  Our idea is to define a discrete neighborhood of the current SGD point containing a number of ``potentially good moves'' that exploit gradient information, and to search this neighborhood by using a classical metaheuristic scheme borrowed from Discrete Optimization. 

In the present paper we investigate the use of a simple Simulated Annealing (SA) metaheuristic that accepts/rejects a candidate new solution in the neighborhood with a probability that depends both on the new solution quality and on a parameter (the \emph{temperature}) which is modified over time to lower the probability of accepting worsening moves. We use this scheme as an automatic way to perform hyper-parameter tuning, hence the title of the paper. A distinctive feature of our scheme is that hyper-parameters are modified within a \emph{single} SGD execution (and not in an external loop, as customary) and evaluated on the fly on the current minibatch, i.e., their tuning is fully embedded within the SGD algorithm.     

The use of SA for training is not new, but previous proposals were mainly intended for non-differentiable objective functions for which SGD is not applied due to the lack of gradients. On the contrary, our SA method requires differentiability of (a proxy of) the loss function, and leverages on the availability of a gradient direction to define local moves that have a large probability to improve the current solution.



Computational results on image classification (CIFAR-10) are reported, showing that the proposed approach leads to an improvement of the final validation accuracy for modern Deep Neural Networks such as ResNet34 and VGG16.

\end{abstract}

\section{Introduction} \label{sec:intro}

Stochastic Gradient Descent (SGD) is \emph{de facto} the standard algorithm for training Deep Neural Networks (DNNs).  Leveraging the gradient, SGD allows one to rapidly find a good solution in the very high dimensional space of weights associated with modern DNNs; moreover, the use of minibatches allows one to exploit modern GPUs and to achieve a considerable computational efficiency.  


It is well known that SGD uses a number of hyper-parameters that are usually very hard to optimize, as they depend on the algorithm and on the underlying dataset. Hyper-parameter search is commonly performed manually, via rules-of-thumb or by testing sets of hyper-parameters on a predefined grid \cite{DBLP:journals/corr/ClaesenM15}. In SGD, momentum \cite{momentum} or Nesterov \cite{nesterov} are widely recognized to increase the speed of convergence. Instead, effective learning rates are highly dependent on DNN architecture and on the dataset of interest \cite{CLR2}, so they are typically selected on a best-practice basis, although methods such as CLR \cite{smith2015cyclical} have been proposed to reduce the number of choices. Finally, \cite{Bergstra:2012:RSH:2188385.2188395} shows empirically and theoretically that randomly chosen trials are more efficient for hyper-parameter optimization than trials on a grid.

In the present paper we investigate the use of an alternative training method borrowed from Mathematical Optimization, namely, the Simulated Annealing (SA) algorithm \cite{Kirkpatrick1983OptimizationBS}. The use of SA for training is not new, but previous proposals are mainly intended to be applied for non-differentiable objective functions for which SGD is not applied due to the lack of gradients; see, e.g., \cite{sa1, sa2}. Instead, our SA method requires differentiability of (a proxy of) the loss function, and leverages on the availability of a gradient direction to define local moves that have a large probability to improve the current solution.

A notable application of our approach is in the context of SGD hyper-parameter tuning---hence the title of the paper. Assume all possible hyper-parameter values (e.g., learning rates for SGD) are collected in a discrete set $H$. At each SGD iteration, we randomly pick one hyper-parameter from $H$, temporarily implement the corresponding \emph{move} as in the classical SGD method (using the gradient information) and evaluate the new point on the current minibatch. If the loss function does not deteriorate too much, we \emph{accept} the move as in the classical SGD method, otherwise we \emph{reject} it: we step back to the previous point, change the minibatch, randomly pick another hyper-parameter from $H$, and repeat. The decision of accepting/rejecting a move is based on the classical SA criterion, and depends of the amount of loss-function worsening and on a certain parameter (the \emph{temperature}) which is  modified over time to lower the probability of accepting worsening moves. 

A distinctive feature of our scheme is that hyper-parameters are modified within a \emph{single} SGD execution (and not in an external loop, as customary) and evaluated on the fly on the current minibatch, i.e., their tuning is fully embedded within the SGD algorithm.     

Computational results are reported, showing that the proposed approach leads to an improvement of the final validation accuracy for modern DNN architectures (ResNet34 and VGG16 on CIFAR-10). Also, it turns out that the random seed used within our algorithm modifies the search path in a very significant way, allowing one to use this seed as a single hyper-parameter to be tuned in an external loop for a further improved generalization.

\section{Simulated Annealing} \label{sec:simulated-annealing}


The basic SA algorithm for a generic optimization problem can be outlined as follows. Let $S$ be the set of all possible feasible solutions, and $f : S \rightarrow \mathbb{R}$ be the objective function to be minimized. An optimal solution $s^{*}$ is a  solution in $S$ such that $f(s^*) \leq f(s)$ holds for all $s \in S$.
                                                                                                             
SA is an iterative method that constructs a trajectory of solutions $s^{(0)}, \cdots, s^{(k)}$ in $S$. At each iteration, SA considers moving from the  current feasible solution $s^{(i)}$ (say) to a candidate new feasible solution $s_{new}$ (say). Let $\Delta(s^{(i)}, s_{new}) = f(s_{new}) - f(s^{(i)})$ be the objective function \emph{worsening} when moving from $s^{(i)}$ to $s_{new}$---positive if $s_{new}$ is strictly worse than $s^{(i)}$.  The hallmark of SA is that worsening moves are not forbidden but accepted with a certain \emph{acceptance probability} $p(s^{(i)},s_{new},T)$ that depends on the amount of worsening $\Delta(s^{(i)}, s_{new})$ and on a parameter $T>0$ called \emph{temperature}. A typical way to compute the acceptance probability is through \emph{Metropolis' formula} \cite{metropolis}:
\begin{equation}                    
	 \label{eq:prob}        
    p(s,s_{new},T)=\left\{\begin{array}{ll}{e^{-\Delta(s^{(i)}, s_{new})/T}} & {\text { if } \quad \Delta(s^{(i)}, s_{new})>0} \\ 
		{1} & {\text { if } \quad \Delta(s^{(i)}, s_{new}) \leq 0 \ .}\end{array}\right.
\end{equation}                                                                     
Thus, the probability of accepting a worsening move is large if the amount of worsening $\Delta(s^{(i)}, s^{\prime})>0$ is small and the temperature $T$ is large. Note that the probability is 1 when $\Delta(s^{(i)}, s^{\prime})\le 0$, meaning that improving moves are always accepted by the SA method. 

Temperature $T$ is a crucial parameter: it is initialized to a certain value $T_0$ (say), and iteratively decreased during the SA execution so as to make worsening moves less and less likely in the final iterations. A simple update formula for $T$ is based on a \emph{cooling factor} $\alpha \in (0,1)$ and reads
\begin{equation}
    T =\alpha \cdot  T \ ;   \label{eq:temperature}
\end{equation}     
typical ranges for $\alpha$  are $0.95-0.99$ (if cooling is applied at each SA iteration) or $0.7-0.8$ (if cooling is only applied at the end of a ``computational epoch'', i.e., after several SA iterations with a constant temperature).    

The basic SA scheme is outlined in Algorithm~\ref{alg:sa}; more advanced implementations are possible, e.g., the temperature can be restored multiple times to the initial value. 


\begin{algorithm}
 \caption{: SA}\label{alg:sa}
 \begin{algorithmic}[1]
  \Statex \textbf{Input:} function $f$ to be minimized, initial temperature $T_0 > 0$, cooling factor $\alpha \in (0, 1)$, number of iterations $nIter$ 
  \Statex \textbf{Output: } the very last solution $s^{(nIter)}$ 
	\vspace{0.1cm}
  \State Compute an initial solution $\mathbf{s}^{(0)}$ and initialize $T = T_0$ 
  \For{$i = 0, \ldots, nIter-1$}
  \State Pick a new tentative solution $s_{new}$ in a convenient neighborhood ${\cal N}(s^{(i)})$ of $s^{(i)}$  
	\State $worsening = f(\mathbf{s}_{new}) - f(\mathbf{s}^{(i)})$
	\State $prob = e^{- worsening/T}$              \label{step:wors}
  \If{ $random(0, 1) < prob$ }                         \label{step:rand}
  \State   $\mathbf { s } ^ {(i + 1)} = \mathbf { s }_{new}$
	\Else
  \State   $\mathbf { s } ^ {(i + 1)} = \mathbf { s } ^ {(i)}$
  \EndIf
  \State $T = \alpha \cdot T$
  \EndFor
 \end{algorithmic}
\end{algorithm}     
At Step~\ref{step:rand}, $random(0,1)$ is a pseudo-random value uniformly distributed in [0,1]. Note that, at Step~\ref{step:wors}, the acceptance probability $prob$ becomes larger than 1 in case $worsening < 0$, meaning that improving moves are always accepted (as required).

\subsection{A naive implementation for training without gradients}
In the context of training, one is interested in minimizing a \emph{loss function} $L(w)$ with respect to a large-dimensional vector $w \in \Re^M$ of so-called \emph{weights}. If $ L ( \mathbf { w } )$ is differentiable (which is not required by the SA algorithm), there exists a gradient $\nabla(\mathbf{w})$                                            
giving the steepest increasing direction of $L$ when moving from a given point $w$. 

Here is a very first attempt to use SA in this setting. Given the current solution (i.e., set of weights) $\mathbf{w}$, we generate a random move $\Delta(\mathbf{w}) \in \Re^M$ and then we evaluate the loss function in the nearby point $\mathbf{w'} := \mathbf{w} - \epsilon \Delta(\mathbf{w})$,  where $\epsilon$ is a small positive real number. If the norm of $\epsilon \Delta(w)$ is small enough and $L$ is differentiable, due to Taylor's approximation we know that 
\begin{equation}
    L(\mathbf{w'}) \simeq L(\mathbf{w}) - \epsilon \ \nabla^T(\mathbf{w}) \Delta(\mathbf{w})  \ .
\end{equation}
Thus the objective function improves if $\nabla(\mathbf{w})^T \Delta(\mathbf{w}) > 0$. As we work in the continuous space, in the attempt of improving the objective function we can also try to move in the opposite direction and move to $\mathbf{w''} := \mathbf{w} + \epsilon \ \Delta(\mathbf{w})$. Thus, our actual move from the current $w$ consists of picking the best (in terms of objective function) point $w_{new}$, say, between the two nearby points $w'$ and $w''$:
if $w_{new}$ improves $L(\mathbf{w})$, then we surely accept this move; otherwise we accept it according to the Metropolis' formula \eqref{eq:prob}.
Note that the above SA approach is completely derivative free: as a matter of fact, SA could optimize directly over discrete functions such as the accuracy in the context of classification.

\begin{figure}
    \centering
    \begin{subfigure}[b]{0.48\textwidth}
        \includegraphics[width=\textwidth]{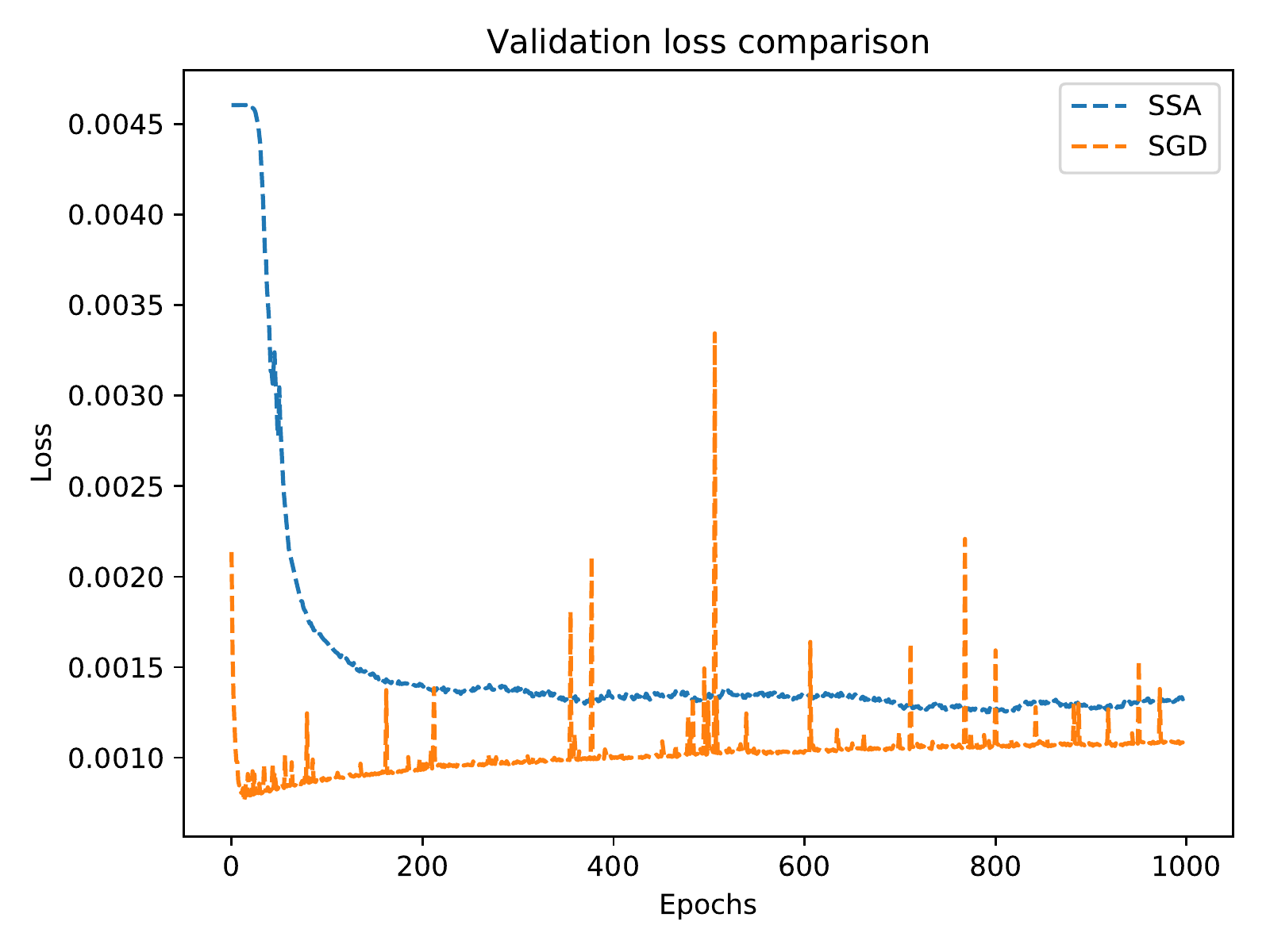}
        \caption{Validation loss}
        \label{fig:vgg16_validation_loss}
    \end{subfigure}
    ~
    \begin{subfigure}[b]{0.48\textwidth}
        \includegraphics[width=\textwidth]{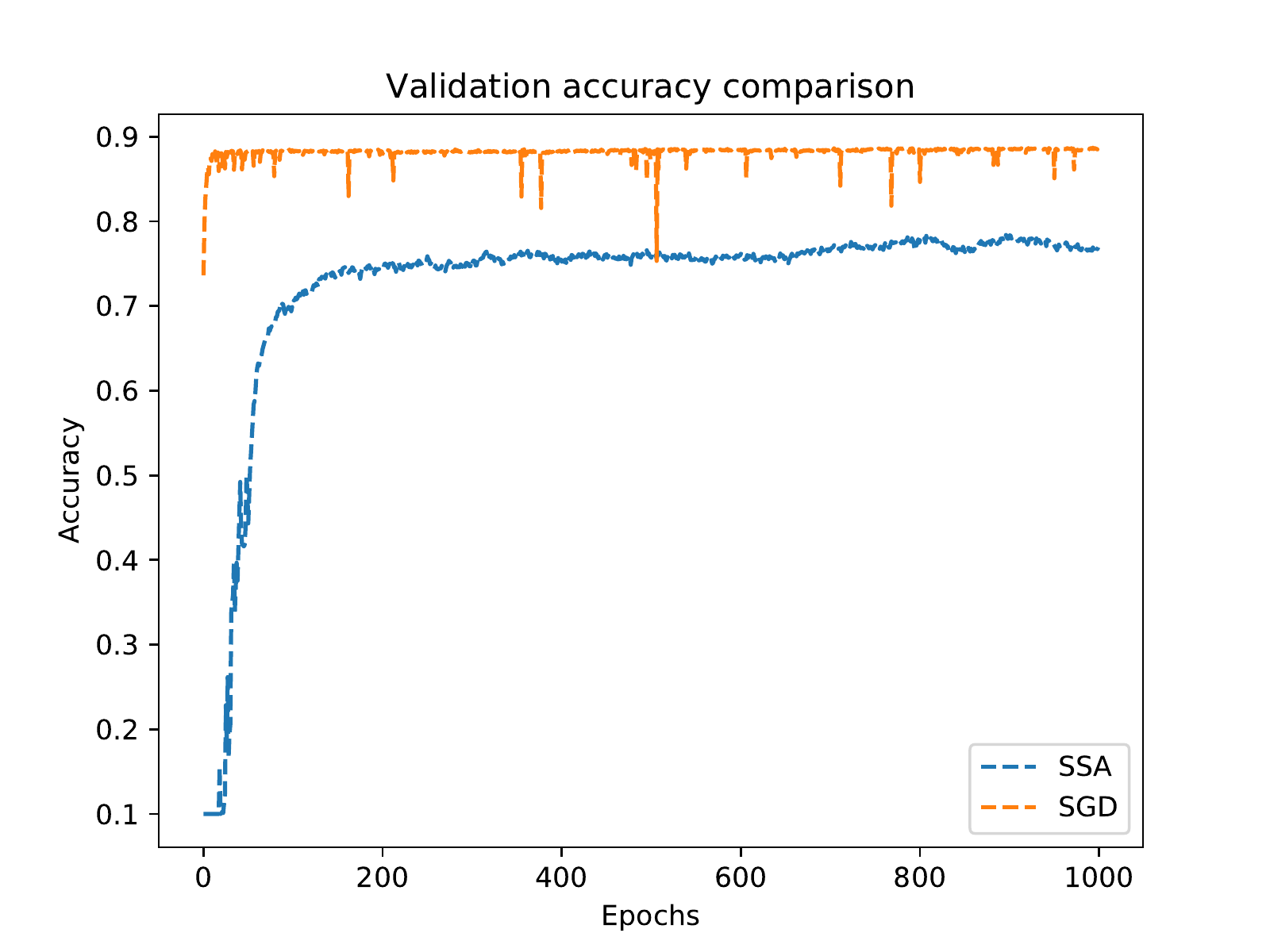}
        \caption{Validation accuracy}
        \label{fig:vgg16_validation_accuracy}
    \end{subfigure}
    
    \begin{subfigure}[b]{0.48\textwidth}
        \includegraphics[width=\textwidth]{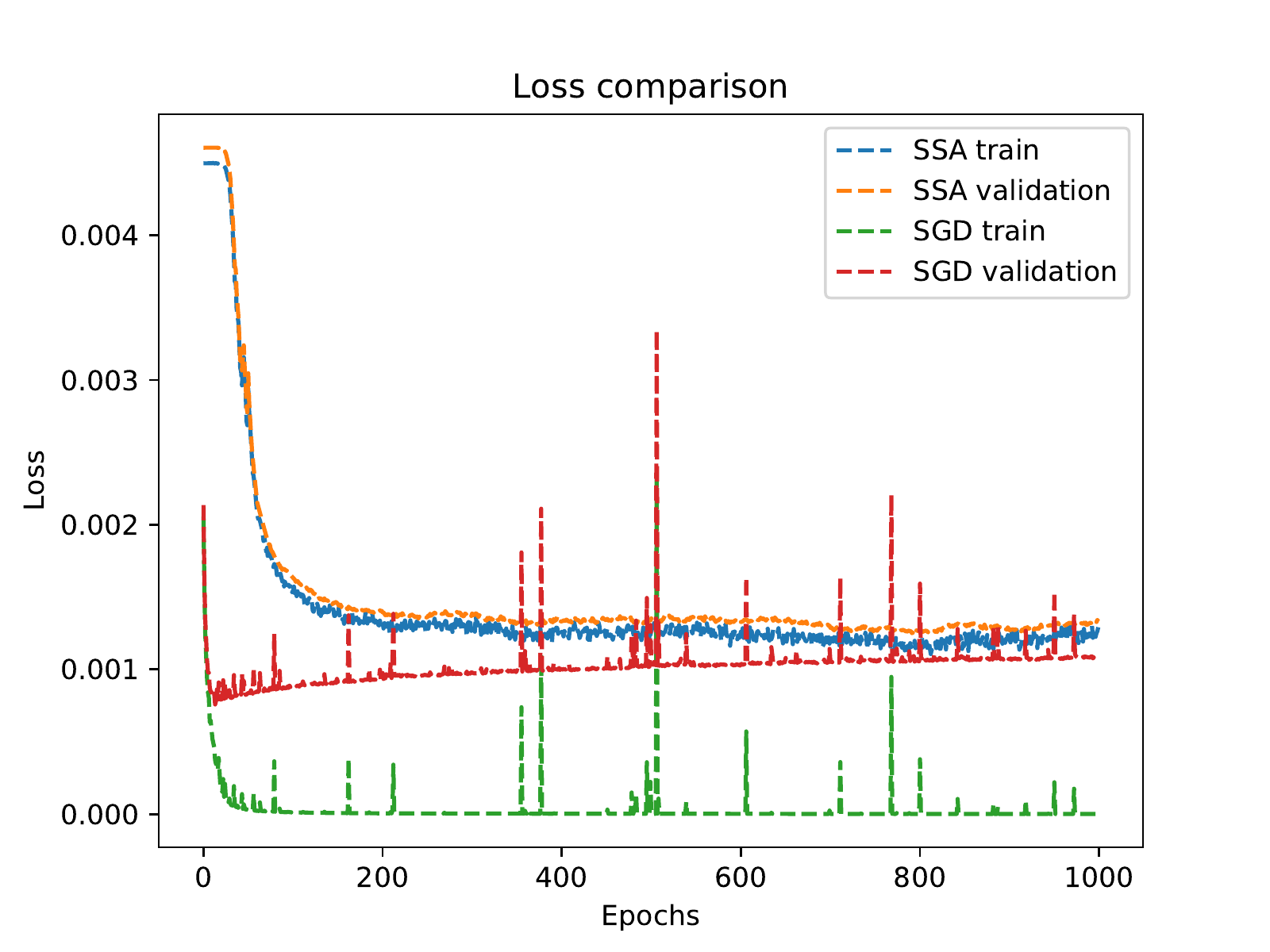}
        \caption{Loss comparison}
        \label{fig:vgg16_loss}
    \end{subfigure}
    ~
    \begin{subfigure}[b]{0.48\textwidth}
        \includegraphics[width=\textwidth]{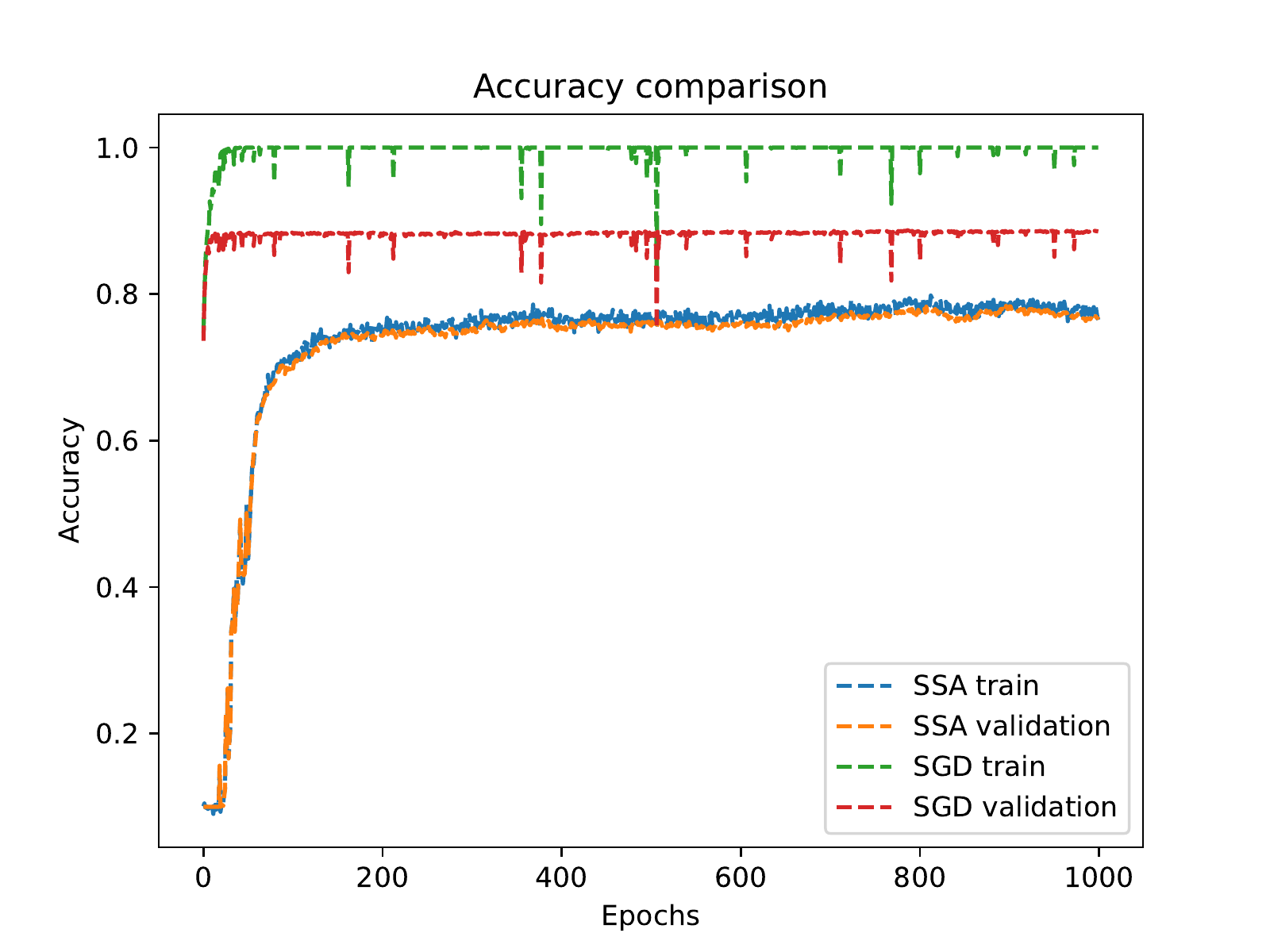}
        \caption{Accuracy comparison}
        \label{fig:vgg16_accuracy}
    \end{subfigure}
    
    \caption{Naive SA implementation for VGG16 on Fashion-MNIST. SGD: learning rate $\eta = 0.001$, no momentum/Nesterov acceleration.  \SSA: $\epsilon = 0.01\text{, }\alpha = 0.97$, $T_0 = 1$. Subfigure (d) clearly shows that \SSA\ has no overfitting but is not able to exploit well the capacity of VGG16, resulting into an unsatisfactory final accuracy.}
    \label{fig:vgg16_ssa}
\end{figure}

In a preliminary work we implemented the simple scheme above in a stochastic manner, using minibatches when evaluating $L(w')$ and $L(w'')$, very much in the spirit of the SGD algorithm. 
\iflong
\else
Figure~\ref{fig:vgg16_ssa} 
\fi
compares the performance of the resulting Stochastic SA algorithm, called \SSA,  with that of a straightforward SGD implementation with constant learning rate and no momentum/Nesterov acceleration, using the Fashion-MNIST \cite{xiao2017/online} dataset and the VGG16 \cite{vgg} architecture. Figure~\ref{fig:vgg16_ssa}(d) reports accuracy on both the training and the validation sets, showing that \SSA\ does not suffer from overfitting as the accuracy on the training and validation sets are almost identical---a benefit deriving from the derivative-free nature of \SSA. However, \SSA\ is clearly unsatisfactory in terms of validation accuracy (which is much worse than the SGD one) in that it does not exploit well the VGG16 capacity.       

We are confident that the above results could be improved by a more advanced implementation. E.g., one could vary the value of $\epsilon$ during the algorithm, and/or replace the loss function by (one minus) the accuracy evaluated on the current minibatch---recall that \SSA\ does not require the objective function be differentiable. However, even an improved \SSA\ implementation is unlikely to be competitive with SGD. In our view, the main drawback of the \SSA\ algorithm (as stated) is that, due the very large dimensional space, the random direction $\pm \Delta(\mathbf{w})$ is very unlikely to lead to a substantial improvement of the objective function as the effect of its components tend to cancel out randomly. Thus, a more clever definition of the basic move is needed to drive \SSA\ in an effective way. 

\section{Improved SGD training by SA}      

We next introduce an unconventional way of using SA in the context of training. We assume the function $L(w)$ to be minimized be differentiable, so we can compute its gradient $\nabla(w)$. From SGD we borrow the idea of moving in the anti-gradient direction $-\nabla(w)$, possibly corrected using momentum/Nesterov acceleration techniques. Instead of using a certain \emph{a priori} learning rate $\eta$, however, we randomly pick one from a discrete set $H$ (say) of possible candidates. In other words, at each SA iteration the move is selected randomly in a discrete neighborhood ${\cal N}(w^{(i)})$ whose elements correspond to SGD iterations with different learning rates. An important feature of our method is that $H$ can (actually, should) contain unusually large learning rates, as the corresponding moves can be discarded by the Metropolis' criterion if they deteriorate the objective function too much. 

A possible interpretation of our approach is in the context of SGD hyper-parameter tuning. According to our proposal, hyper-parameters are collected in a discrete set $H$ and sampled \emph{within a single SGD execution}: in our tests, $H$ just contains a number of possible learning rates, but it could involve other parameters/decisions as well, e.g., applying momentum, or Nesterov (or none of the two) at the current SGD iteration, or alike. The key property here is that any element in $H$ corresponds to a reasonable (non completely random) move, so picking one of them at random has a significant probability of improving the objective function. As usual, moves are accepted according to the Metropolis' criterion, so the set $H$ can also contain ``risky choices'' that would be highly inefficient if applied systematically within a whole training epoch.

\begin{algorithm}
 \caption{: \NOI}\label{alg:sgd_sa}
 \begin{algorithmic}[1]
  \Statex \textbf{Parameters:} A set of learning rates $H$, initial temperature $T_0 > 0$
  \Statex \textbf{Input:} Differentiable loss function $L$ to be minimized, cooling factor $\alpha \in (0, 1)$, number of epochs $nEpochs$, number of minibatches $N$
  \Statex \textbf{Output: } the best performing $\mathrm { \mathbf{w} } ^ { ( i ) }$ on the validation set at the end of each epoch  
	\vspace{0.1cm}
	\State Divide the training dataset into $N$ minibatches
  \State Initialize $i = 0$, \ $T = T_0$, \ $\mathbf{w}^{(0)}$ = random\_initialization()     \label{step:init}
  \For{$t = 1, \ldots, nEpochs$}
  \For{$n = 1, \ldots, N$}
  \State Extract the $n$-th minibatch ($\mathbf{x}$, $\mathbf{y}$)
  \State Compute $L(\mathbf{w}^{(i)}, \mathbf{x}, \mathbf{y})$ and its gradient $\mathbf{ v } = \text { backpropagation}(\mathbf {w}^{(i)},\mathbf{ x }, \mathbf { y })$
  \State Randomly pick a learning rate $\eta$ from $H$    \label{step:eta}
  \State $\mathbf { w }_{new} = \mathbf { w } ^ { ( i ) } - \eta \ \mathbf { v }$ 
  \State Compute $L(\mathbf{w}_{new}, \mathbf{x}, \mathbf{y})$
	\State ${worsening} = L(\mathbf{w}_{new}, \mathbf{x}, \mathbf{y}) - L(\mathbf{w}^{(i)}, \mathbf{x}, \mathbf{y})$
  \State ${prob} = {e^{- {worsening}/T}}$
  \If{ $random(0, 1) < prob$ }          \label{step:accept}       
  \State   $\mathbf { w } ^ {(i + 1)} = \mathbf { w }_{new}$
	\Else
  \State   $\mathbf { w } ^ {(i + 1)} = \mathbf { w } ^ {(i)}$
  \EndIf
  \State $i = i + 1$
  \EndFor
  \State $T = \alpha \cdot T$
  \EndFor
 \end{algorithmic}
\end{algorithm}

Our basic approach is formalized in Algorithm \ref{alg:sgd_sa}, and will be later referred to as \NOI. More elaborated versions using momentum/Nesterov are also possible but not investigated in the present paper, as we aim at keeping the overall computational setting as simple and clean as possible.

\section{Computational analysis of \NOI} \label{sec:computational-analysis}   

We next report a computational comparison of SGD and \NOI\ for a classical image classification task involving the CIFAR-10 \cite{cifar10} dataset. As customary, the dataset was shuffled  and partitioned into 50,000 examples for the training set, and the remaining 10,000 for the test set. As to the DNN architecture, we tested two well-known proposals from the literature: VGG16 \cite{vgg} and ResNet34 \cite{resnet}. Training was performed for 100 epochs using PyTorch, with minibatch size 512. Tests have been performed using a single NVIDIA TITAN Xp GPU. 

Our \SGD\ implementation of SGD is quite basic but still rather effective on our dataset: it uses no momentum/Nesterov acceleration, and the learning rate is set according the following schedule: $\eta = 0.1$ for first $30$ epochs, $0.01$ for the next $40$ epochs, and $0.001$ for the final 30 epochs. As to \NOI, we used $\alpha=0.8$, 
initial temperature $T_0=1$, and learning-rate set $H=\{0.9, 0.8, 0.7, 0.6, 0.5, 0.4, 0.3, 0.2, 0.1, 0.09, 0.08, 0.07, 0.06, 0.05\}$.                                                                                               
              
Both \SGD\ and \NOI\ use pseudo-random numbers generated from an initial random seed, which therefore has some effects of the search path in the weight space and hence on the final solution found. Due to the very large number of weights that lead to statistical compensation effects, the impact of the seed on the initialization of the very first solution $w^{(0)}$ is very limited---a property already known for SGD that is inherited by \NOI\ as well. However, random numbers are used by \NOI\ also when taking some crucial ``discrete'' decisions, namely: the selection of the learning rate $\eta \in H$ (Step~\ref{step:eta}) and the acceptance test (Step~\ref{step:accept}). As a result, as shown next, the search path of \NOI\ is very dependent on the initial seed. Therefore, for both \SGD\ and \NOI\ we decided to repeat each run 10 times, starting with 10 random seeds, and to report results for each seed. In our view, this dependency on the seed is in fact a \emph{positive} feature of \NOI, in that it allows one to treat the seed as a single (quite powerful) hyper-parameter to be randomly tuned in an external loop.

\begin{figure}[h!]
 \centering
 \begin{subfigure}[b]{0.48\textwidth}
 \includegraphics[width=\textwidth]{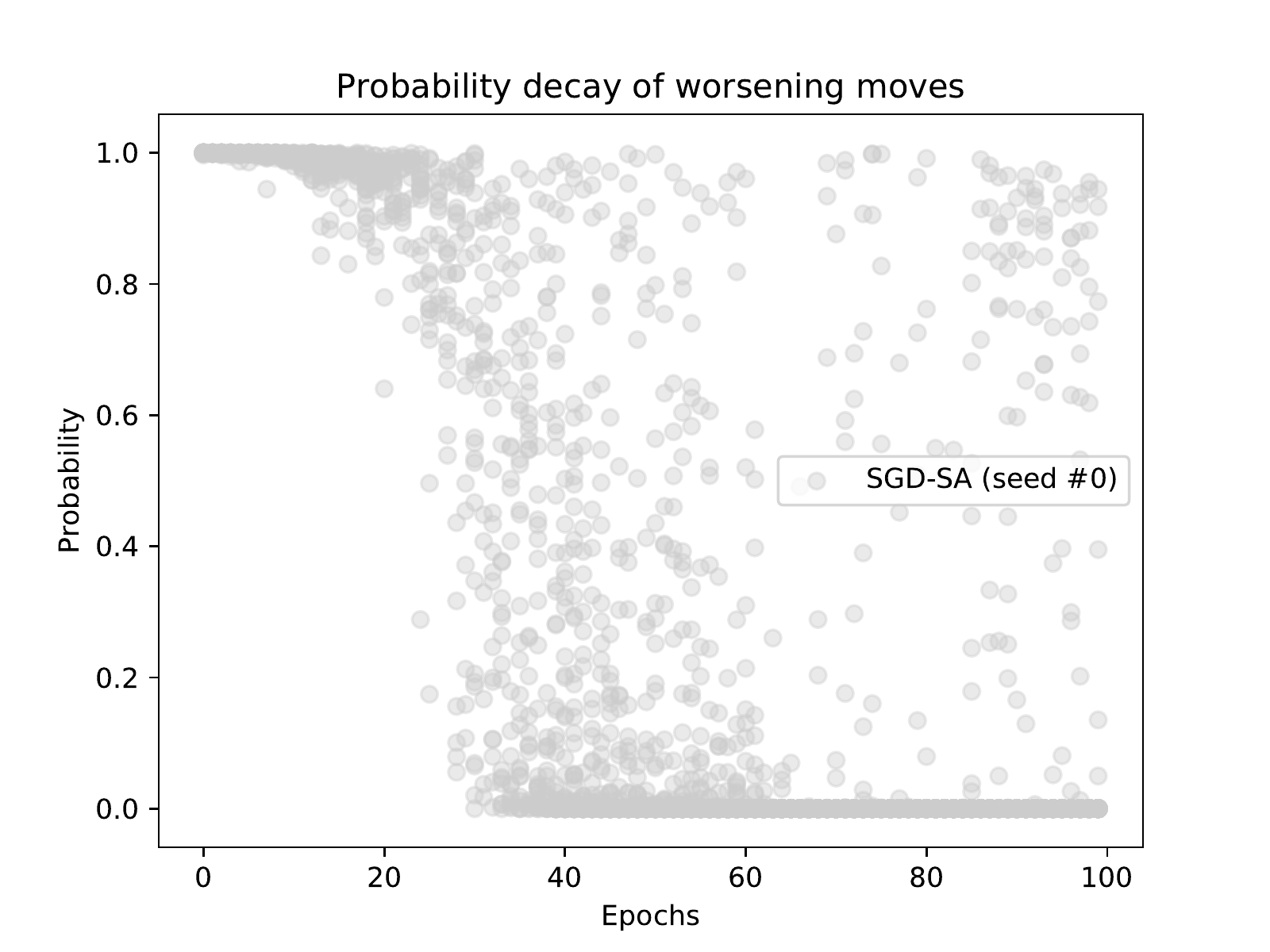}
 \caption{Probability of accepting worsening moves} 
 \label{img:probabilities}
 \end{subfigure}
~
 \begin{subfigure}[b]{0.48\textwidth}
 \includegraphics[width=\textwidth]{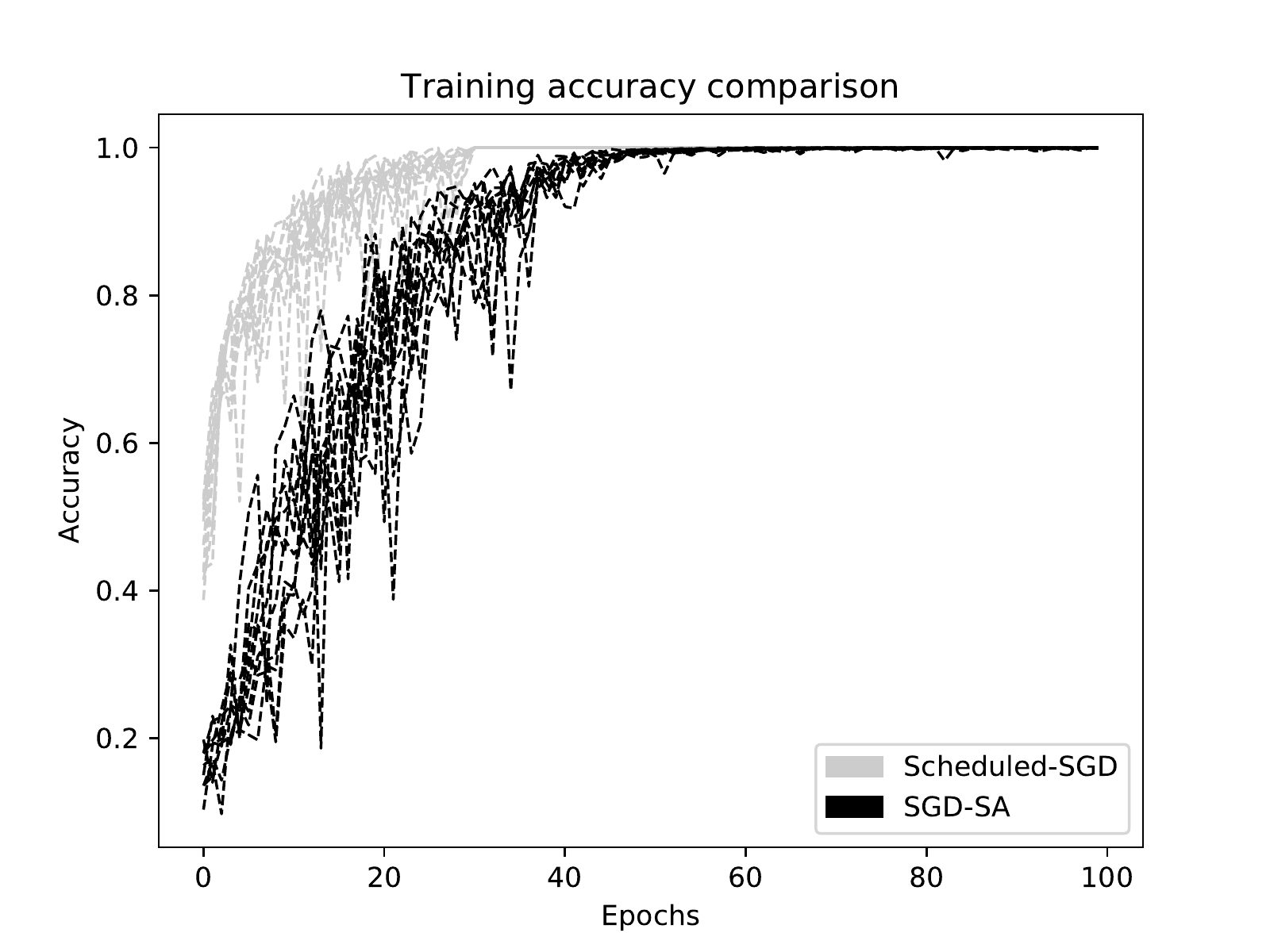}
 \caption{Training accuracy}
 \label{img:train_accuracy}
 \end{subfigure}
 \caption{Optimization efficiency over the training set (VGG16 on CIFAR-10)}   
 \label{img:training} 
\end{figure}

Our first order of business is to evaluate the convergence property of \NOI\ on the training set---after all, this is the optimization task that SA faces directly. In Figure~\ref{img:training} we plot the average probability $prob$ (clipped to 1) of accepting a move at Step~\ref{step:accept}, as well as the training-set accuracy as a function of the epochs. Subfigure \ref{img:probabilities} shows that the probability of accepting a move is almost one in the first epochs, even if the amount of worsening is typically quite large in this phase. Later on, the probability becomes smaller and smaller, and only very small worsenings are more likely to be accepted. As a result, large learning rates are automatically discarded in the last iterations. Subfigure \ref{img:train_accuracy} is quite interesting: even in our simple implementation, \SGD\ quickly converges to the best-possible value of one for accuracy, and the plots for the various seeds (gray lines) are almost overlapping---thus confirming that the random seed has negligible effects of \SGD. As to \NOI\ (black lines), its convergence to accuracy one is slower than \SGD, and different seeds lead to substantially different curves---a consequence of the discrete random decisions taken along the search path. 

\begin{figure}[H]
 \centering
 \begin{subfigure}[b]{0.48\textwidth}
  \includegraphics[width=\textwidth]{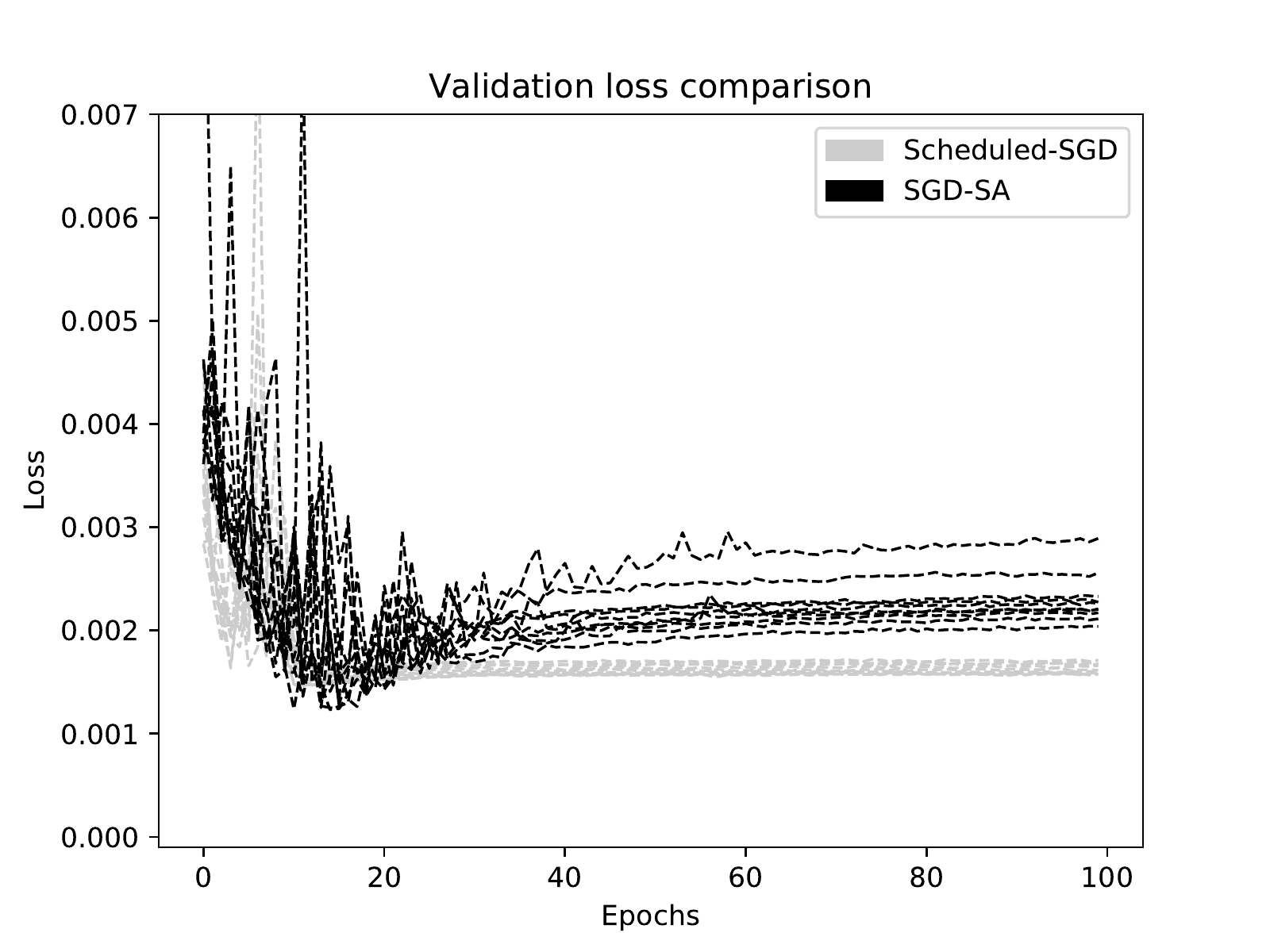}
  \caption{Validation loss}
  \label{fig:resnet_validation_loss}
 \end{subfigure}
 ~
 \begin{subfigure}[b]{0.48\textwidth}
  \includegraphics[width=\textwidth]{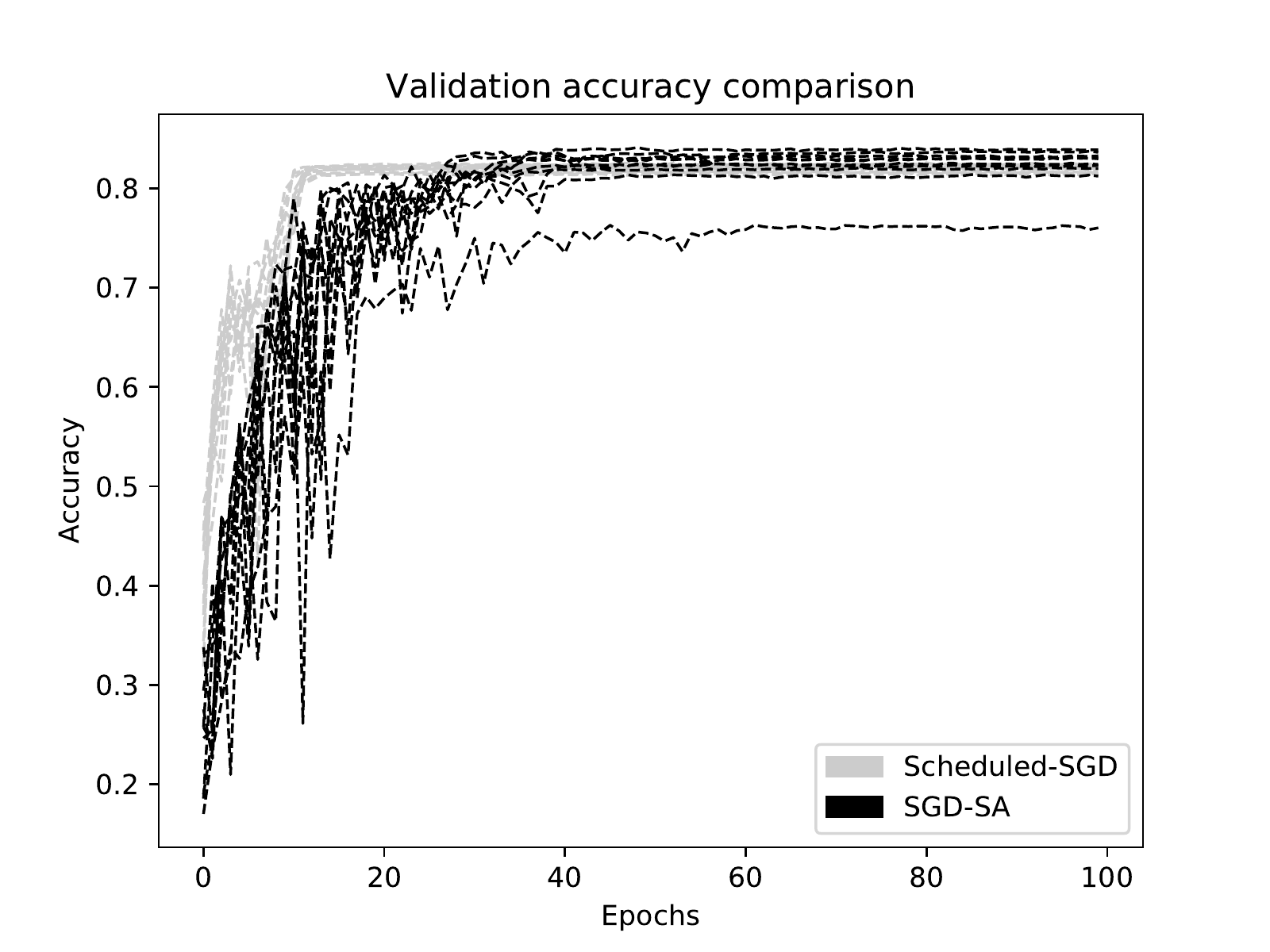}
  \caption{Validation accuracy}
  \label{fig:resnet_validation_accuracy}
 \end{subfigure}

 \begin{subfigure}[b]{0.48\textwidth}
  \includegraphics[width=\textwidth]{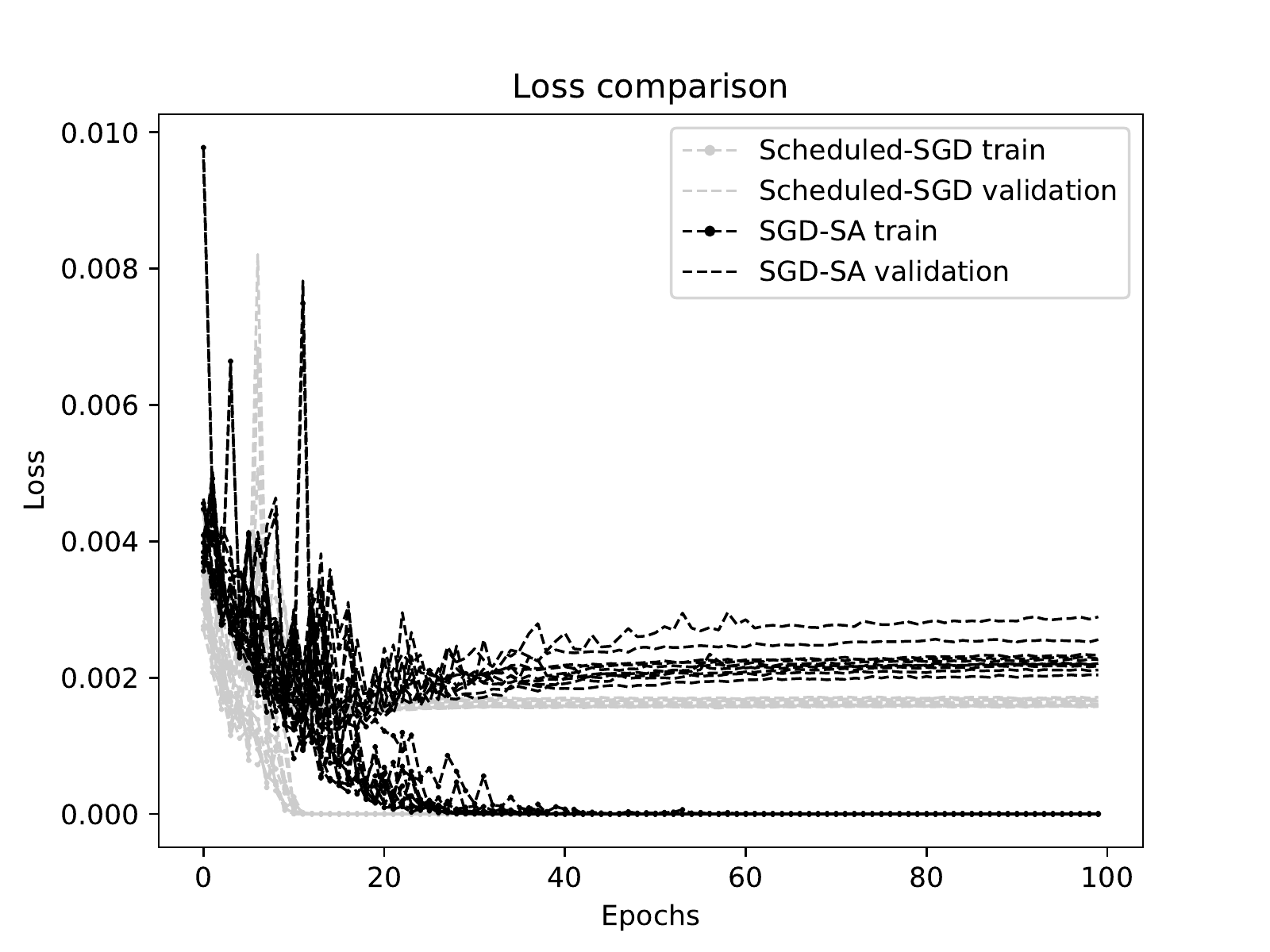}
  \caption{Loss comparison}
  \label{fig:resnet_loss}
 \end{subfigure}
 ~
 \begin{subfigure}[b]{0.48\textwidth}
  \includegraphics[width=\textwidth]{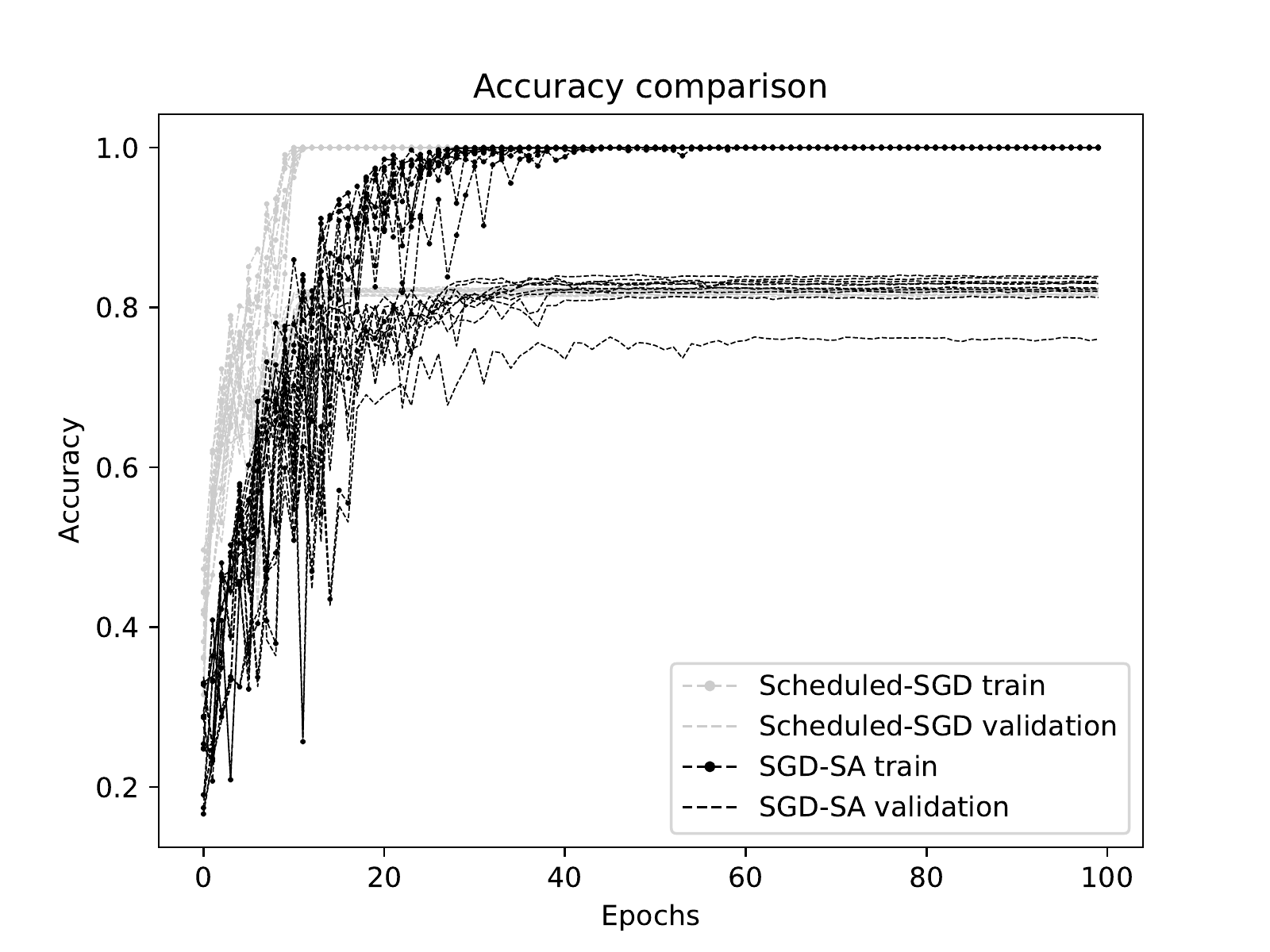}
  \caption{Accuracy comparison}
  \label{fig:resnet_set2_accuracy}
 \end{subfigure}

 \caption{ResNet34 on CIFAR-10 }
 \label{fig:resnet}
\end{figure}

\begin{figure}[H]
 \centering
 \begin{subfigure}[b]{0.48\textwidth}
  \includegraphics[width=\textwidth]{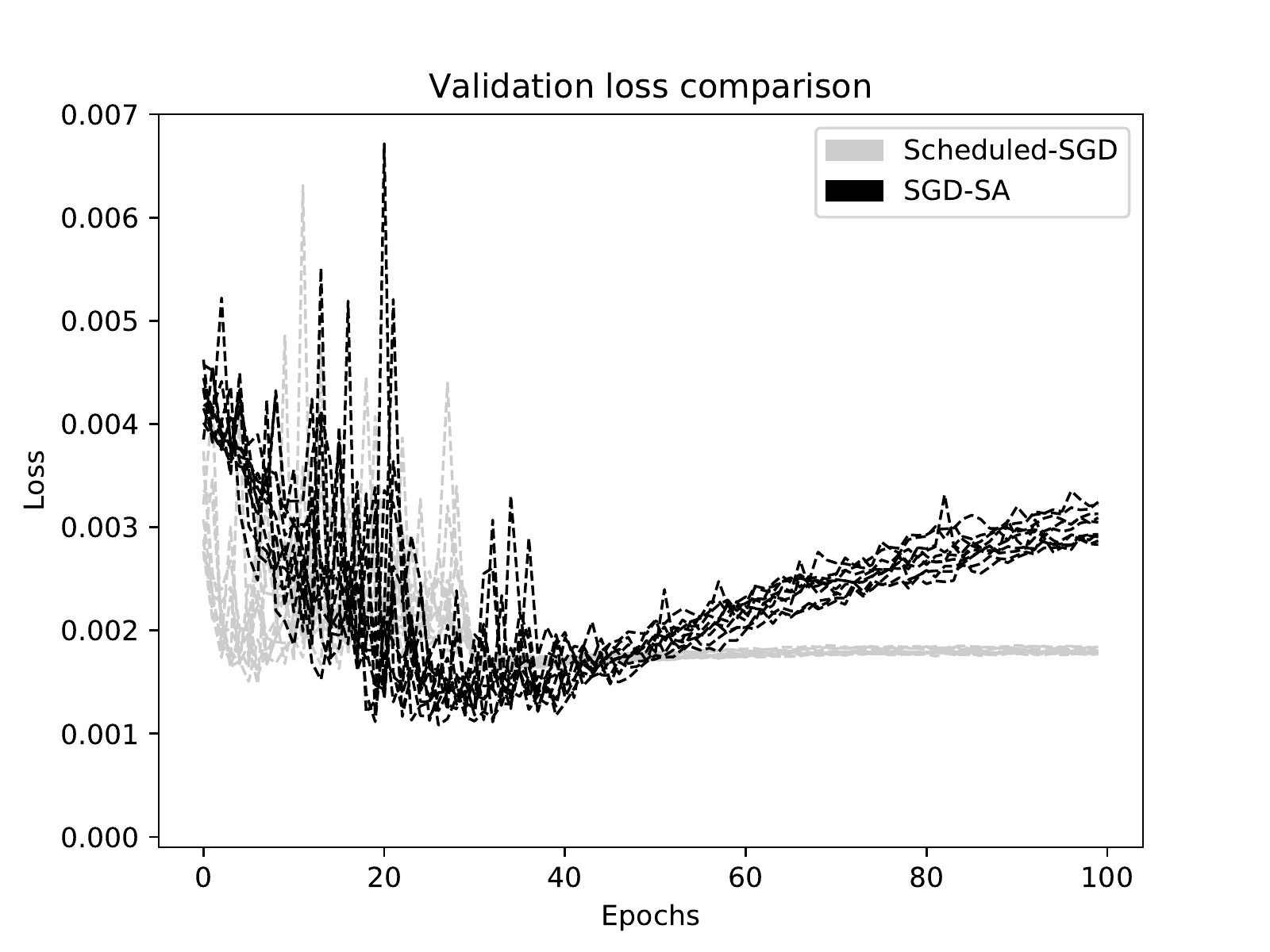}
  \caption{Validation loss}
  \label{fig:vgg_validation_loss}
 \end{subfigure}
 ~
 \begin{subfigure}[b]{0.48\textwidth}
  \includegraphics[width=\textwidth]{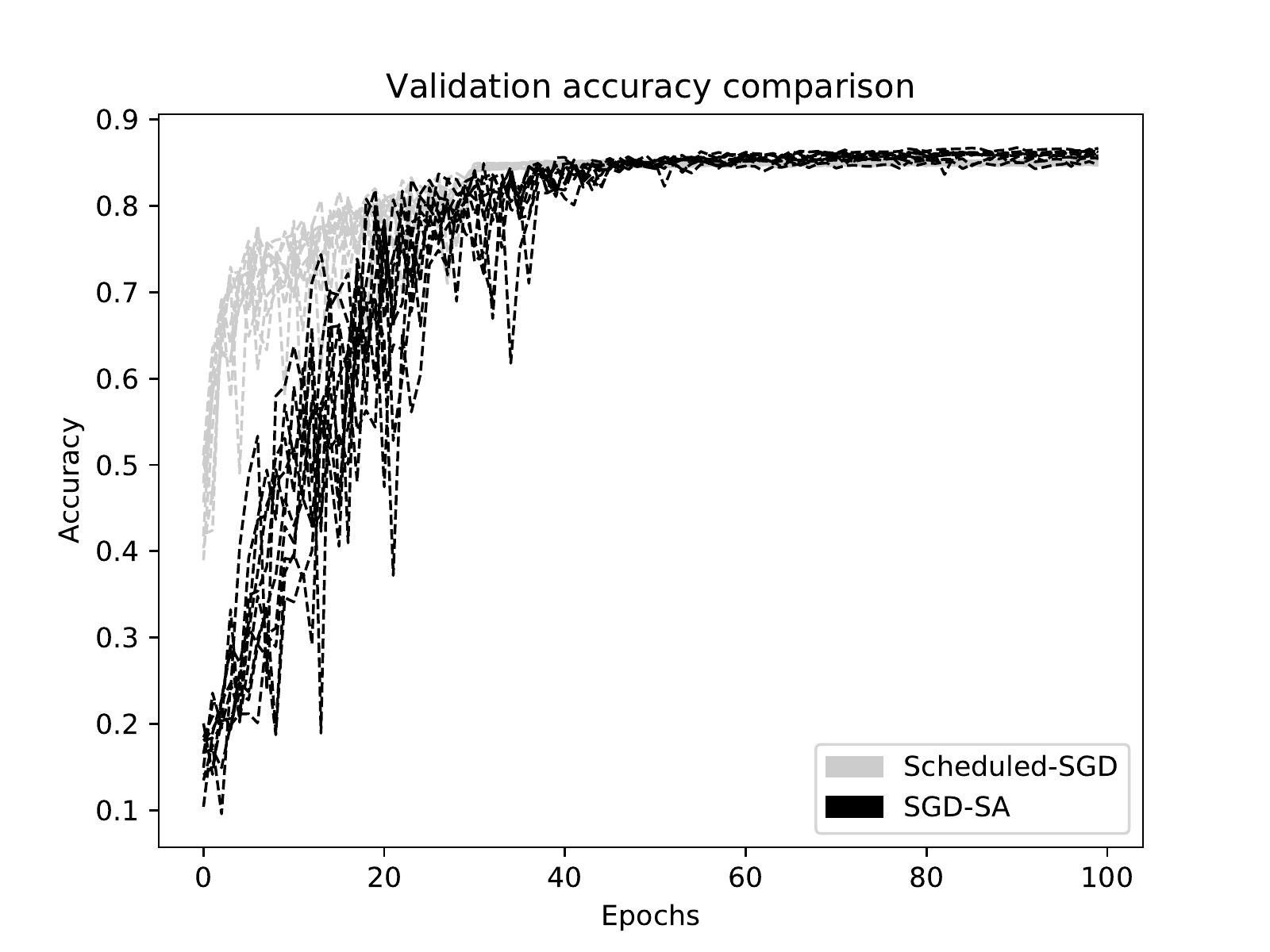}
  \caption{Validation accuracy}
  \label{fig:vgg_validation_accuracy}
 \end{subfigure}

 \begin{subfigure}[b]{0.48\textwidth}
  \includegraphics[width=\textwidth]{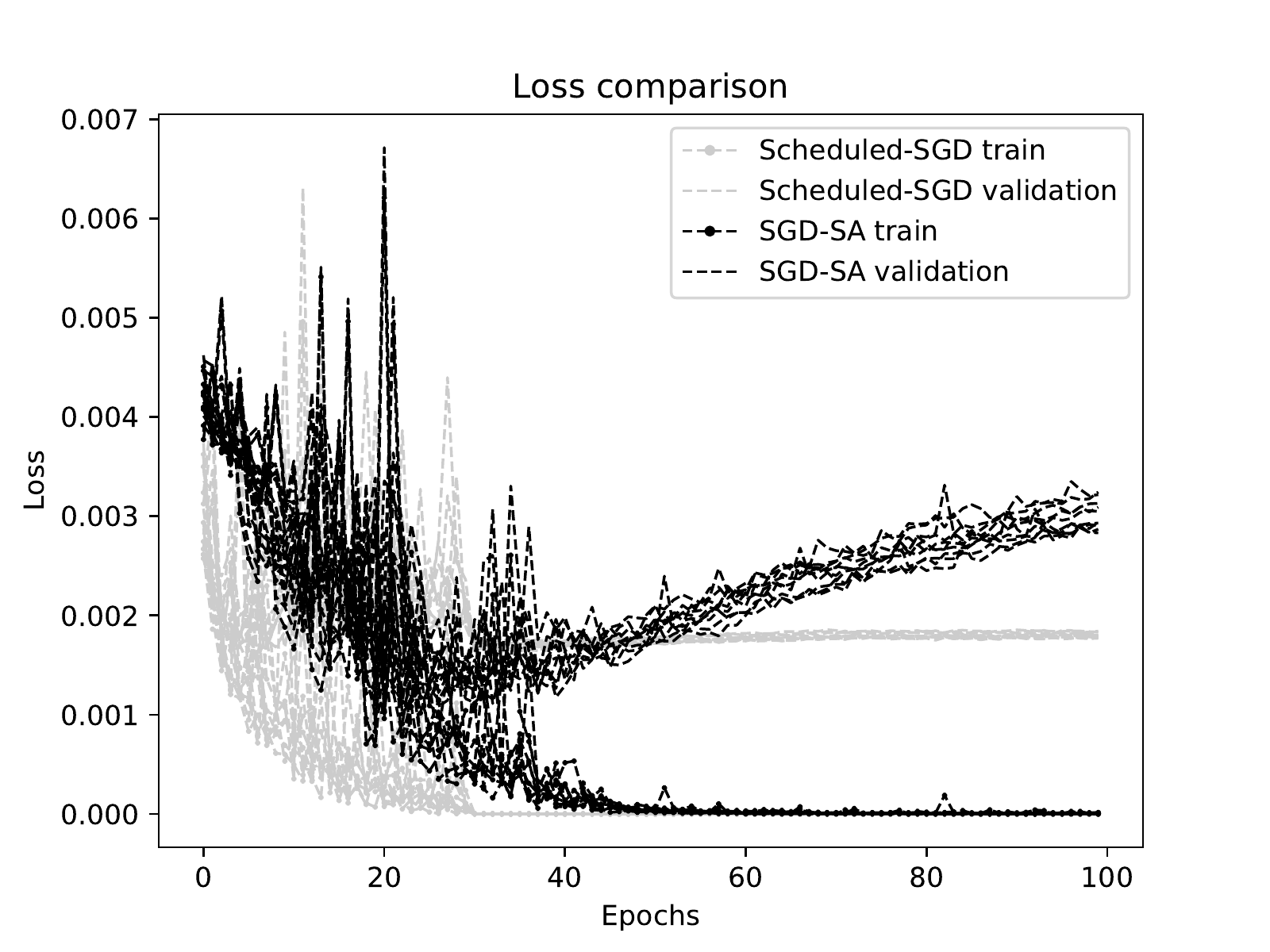}
  \caption{Loss comparison}
  \label{fig:vgg_loss}
 \end{subfigure}
 ~
 \begin{subfigure}[b]{0.48\textwidth}
  \includegraphics[width=\textwidth]{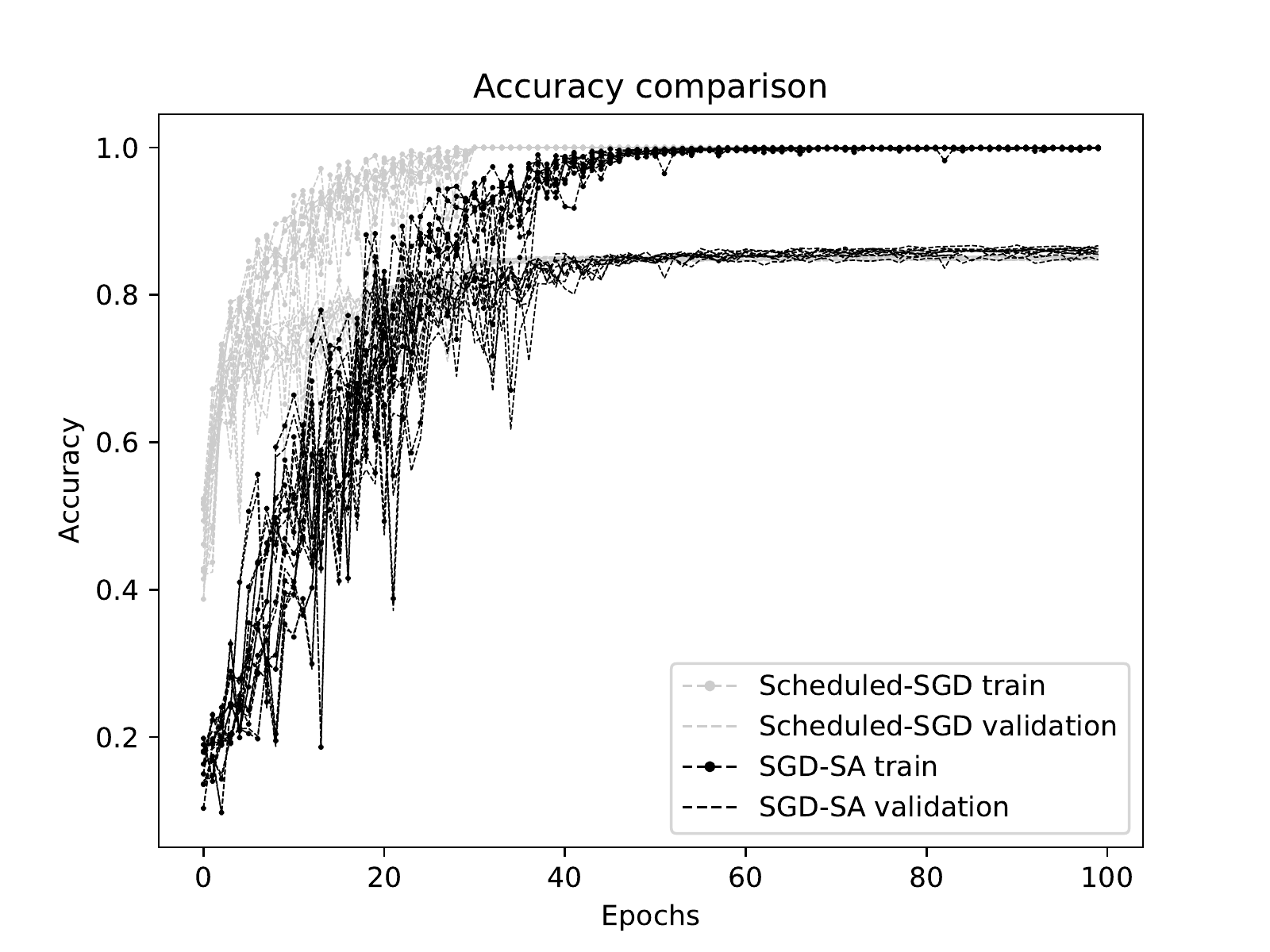}
  \caption{Accuracy comparison}
  \label{fig:vgg_set2_accuracy}
 \end{subfigure}

 \caption{VGG16 on CIFAR-10 } 
 \label{fig:vgg}
\end{figure}

Plots in Figures~\ref{fig:resnet} and \ref{fig:vgg} show the performance (on both the training and validation sets) of \SGD\ and \NOI\ when using the ResNet34 and VGG16 architectures, respectively. As expected, the search path of \NOI\ is more diversified (leading to accuracy drops in the first epochs) abut the final solutions tend to generalize better than \SGD\ (as witnessed by the performance on the validation set).  

Table~\ref{tab:sgd_sa2} gives more detailed results for each seed, and reports the final validation accuracy and loss reached by \SGD\ and \NOI. The results show that, for all seeds, \NOI\ always produces a significantly better (lower) validation loss than \SGD. As to validation accuracy, \NOI\ outperforms \SGD\ for all seeds but seeds 3, 4 and 6 for ResNet34. In particular, \NOI\ leads to a significantly better (1-2\%) validation accuracy than \SGD\ if the best run for the 10 seeds is considered.     
  
\begin{table}[h!]
 \centering
 \footnotesize
 \begin{filecontents*}{csv_example2.csv}
index,seed,method,vggloss,vggacc,resnetloss,resnetacc
0,0,\multirow{10}{*}{Scheduled-SGD},0.001640,85.27,0.001519,82.18
1,1,,0.001564,84.94,0.001472,82.58
2,2,,0.001642,84.84,0.001467,82.27
3,3,,0.001662,84.93,0.001468,82.37
4,4,,0.001628,84.92,0.001602,81.69
5,5,,0.001677,85.37,0.001558,81.80
6,6,,0.001505,84.91,0.001480,82.24
7,7,,0.001480,85.28,0.001532,82.07
8,8,,0.001623,85.26,0.001574,81.52
9,9,,0.001680,85.41,0.001499,82.41 \\ \cline{1-1}
10,0,\multirow{10}{*}{\NOI},0.001127,86.44,0.001306,82.55
11,1,,0.001206,86.18,0.001231,84.11
12,2,,0.001121,86.04,0.001238,83.32
13,3,,0.001133,86.76,0.001457,81.39
14,4,,0.001278,85.17,0.001585,76.31
15,5,,0.001112,86.30,0.001276,83.74
16,6,,0.001233,85.71,0.001405,82.07
17,7,,0.001130,86.59,0.001261,82.57
18,8,,0.001167,86.14,0.001407,83.12
19,9,,0.001084,86.28,0.001240,83.19 \\ \hline
20,,Best \SGD,0.001480,85.41,0.001467,82.58
21,,Best \NOI,\textbf{0.001084},\textbf{86.76},\textbf{0.001240},\textbf{84.11}
\end{filecontents*}
\csvreader[head to column names,
  tabular=lllrlr,
  table head=\toprule
  Method & Seed & \multicolumn{2}{c}{VGG16} & \multicolumn{2}{c}{ResNet34} \\
  \cmidrule(lr){3-4} \cmidrule(lr){5-6}
  & & Loss & Accuracy & Loss & Accuracy \\
  \midrule,
  table foot=\bottomrule,
  after reading = \vspace{5pt}
 ]{csv_example2.csv}{}{\method & \seed & \vggloss & \vggacc & \resnetloss & \resnetacc}
 \caption{Final validation accuracy and loss, seed by seed.}
 \label{tab:sgd_sa2}
\end{table}

\section{Conclusions and future work} \label{sec:conclusions}

We have proposed a new metaheuristic training scheme that combines Stochastic Gradient Descent and Discrete Optimization in an unconventional way. 

Our idea is to define a discrete neighborhood of the current solution containing a number of ``potentially good moves'' that exploit gradient information, and to search this neighborhood by using a classical metaheuristic scheme borrowed from Discrete Optimization. In the present paper, we have investigated the use of a simple Simulated Annealing metaheuristic that accepts/rejects a candidate new solution in the neighborhood with a probability that depends both on the new solution quality and on a parameter (the temperature) which is varied over time. We have used this scheme as an automatic way to perform hyper-parameter tuning within a single training execution, and have shown its potentials on a classical test problem (CIFAR-10 image classification using VGG16/ResNet34 deep neural networks).
                                        
In a follow-up research we plan to investigate the use of two different objective functions at training time: one differentiable to compute the gradient (and hence a set of potentially good moves), and one completely generic (possibly black-box) for the Simulated Annealing acceptance/rejection test---the latter intended to favor simple/robust solutions that are likely to generalize well.

Replacing Simulated Annealing with other Discrete Optimization metaheuristics (tabu search, variable neighborhood search, genetic algorithms, etc.) is also an interesting topic that deserves future research.


\section*{Acknowledgments}
Work supported by MiUR, Italy (project PRIN 2015 on ``Nonlinear and Combinatorial Aspects of Complex Networks''). We gratefully acknowledge the support of NVIDIA Corporation with the donation of the Titan Xp GPU used for this research.


\bibliographystyle{plain}       
\bibliography{sgd-sa}

\begin{thebibliography}{10}

\bibitem{Bergstra:2012:RSH:2188385.2188395}
James Bergstra and Yoshua Bengio.
\newblock Random search for hyper-parameter optimization.
\newblock {\em J. Mach. Learn. Res.}, 13:281--305, February 2012.

\bibitem{DBLP:journals/corr/ClaesenM15}
Marc Claesen and Bart~De Moor.
\newblock Hyperparameter search in machine learning.
\newblock {\em CoRR}, abs/1502.02127, 2015.

\bibitem{resnet}
K.~{He}, X.~{Zhang}, S.~{Ren}, and J.~{Sun}.
\newblock {Deep Residual Learning for Image Recognition}.
\newblock {\em ArXiv e-prints}, December 2015.

\bibitem{Kirkpatrick1983OptimizationBS}
Scott Kirkpatrick, C.~D. Gelatt, and Mario~P. Vecchi.
\newblock Optimization by simulated annealing.
\newblock {\em Science}, 220 4598:671--80, 1983.

\bibitem{cifar10}
Alex Krizhevsky, Vinod Nair, and Geoffrey Hinton.
\newblock {CIFAR-10 (Canadian Institute for Advanced Research)}.
\newblock {\em {\tt http://www.cs.toronto.edu/\textasciitilde
  kriz/cifar.html}}.

\bibitem{sa2}
Sergio Ledesma, Miguel Torres, Donato Hern{\'a}ndez, Gabriel Avi{\~{n}}a, and
  Guadalupe Garc{\'i}a.
\newblock Temperature cycling on simulated annealing for neural network
  learning.
\newblock In Alexander Gelbukh and {\'A}ngel~Fernando Kuri~Morales, editors,
  {\em MICAI 2007: Advances in Artificial Intelligence}, pages 161--171,
  Berlin, Heidelberg, 2007. Springer Berlin Heidelberg.

\bibitem{metropolis}
Nicholas~Constantine Metropolis, Arianna~W. Rosenbluth, Marshall~N. Rosenbluth,
  Augusta~H. Teller, and Edward Teller.
\newblock Equation of state calculation by fast computing machines.
\newblock {\em Journal of Chemical Physics}, 21:1087--1092, 1953.

\bibitem{nesterov}
Yurii Nesterov.
\newblock {A method of solving a convex programming problem with convergence
  rate O(1/sqr(k))}.
\newblock {\em Soviet Mathematics Doklady}, 27:372--376, 1983.

\bibitem{sa1}
Randall Sexton, Robert Dorsey, and John Johnson.
\newblock Beyond backpropagation: Using simulated annealing for training neural
  networks.
\newblock {\em Journal of End User Computing}, 11, 07 1999.

\bibitem{vgg}
K.~{Simonyan} and A.~{Zisserman}.
\newblock {Very Deep Convolutional Networks for Large-Scale Image Recognition}.
\newblock {\em ArXiv e-prints}, September 2014.

\bibitem{smith2015cyclical}
Leslie~N. Smith.
\newblock Cyclical learning rates for training neural networks, 2015.
\newblock cite arxiv:1506.01186Comment: Presented at WACV 2017; see
  https://github.com/bckenstler/CLR for instructions to implement CLR in Keras.

\bibitem{CLR2}
Leslie~N. Smith.
\newblock A disciplined approach to neural network hyper-parameters: Part 1 -
  learning rate, batch size, momentum, and weight decay.
\newblock {\em CoRR}, abs/1803.09820, 2018.

\bibitem{momentum}
Ilya Sutskever, James Martens, George Dahl, and Geoffrey Hinton.
\newblock On the importance of initialization and momentum in deep learning.
\newblock In {\em Proceedings of the 30th International Conference on
  International Conference on Machine Learning - Volume 28}, ICML'13, pages
  III--1139--III--1147. JMLR.org, 2013.

\bibitem{xiao2017/online}
Han Xiao, Kashif Rasul, and Roland Vollgraf.
\newblock Fashion-mnist: a novel image dataset for benchmarking machine
  learning algorithms, 2017.

\end{thebibliography}

\end{document}